\documentclass[conference]{IEEEtran}
\IEEEoverridecommandlockouts
\usepackage{cite}
\usepackage{amsmath,amssymb,amsfonts}
\usepackage{graphicx}
\usepackage{textcomp}
\usepackage{xcolor}
\usepackage{amsmath}
\usepackage{graphicx}
\usepackage{booktabs}
\usepackage{multirow}
\usepackage{caption}
\usepackage{subcaption}
\usepackage{url}
\usepackage{algorithm}
\usepackage{algpseudocode}
\usepackage{tikz}
\usepackage{pgfplots}
\usepackage{enumitem}
\newcommand{\frameworkname}{Event-CausNet}

\def\BibTeX{{\rm B\kern-.05em{\sc i\kern-.025em b}\kern-.08em
    T\kern-.1667em\lower.7ex\hbox{E}\kern-.125emX}}
\begin{document}

\title{Event-CausNet: Unlocking Causal Knowledge from Text with Large Language Models for Reliable Spatio-Temporal Forecasting
%Beyond Correlations: A Causal Reasoning Framework for Interpretable and Reliable Traffic Forecasting
}

\author{
    \IEEEauthorblockN{
        Luyao Niu\IEEEauthorrefmark{1}\IEEEauthorrefmark{2},
        Zepu Wang\IEEEauthorrefmark{1},
        Shuyi Guan,
        Yang Liu\IEEEauthorrefmark{3},
        and Peng Sun\IEEEauthorrefmark{1}\IEEEauthorrefmark{4}\textsuperscript{,\#}
    }
    \vspace{1em}
    \IEEEauthorblockA{\IEEEauthorrefmark{1} Duke Kunshan University, China}
    \IEEEauthorblockA{\IEEEauthorrefmark{2} Peking University, China}
    \IEEEauthorblockA{\IEEEauthorrefmark{3} Tongji University, China}
    \IEEEauthorblockA{\IEEEauthorrefmark{4} University of Ottawa, Canada}
    \thanks{\# Corresponding author. Email: peng.sun568@duke.edu.
    
    % This work is supported by the Kunshan Municipal Special Project under Grant No. xxxx, the Professional Discretionary Fund xxx, and the Mitacs Elevate Canada under Grant xxx.
    }
}

\maketitle

\begin{abstract}
While spatio-temporal Graph Neural Networks (GNNs) excel at modeling recurring traffic patterns, their reliability plummets during non-recurring events like accidents. This failure occurs because GNNs are fundamentally correlational models, learning historical patterns that are invalidated by the new causal factors introduced during disruptions. To address this, we propose \frameworkname{}, a framework that uses a Large Language Model to quantify unstructured event reports, builds a causal knowledge base by estimating average treatment effects, and injects this knowledge into a dual-stream GNN-LSTM network using a novel causal attention mechanism to adjust and enhance the forecast. Experiments on a real-world dataset demonstrate that \frameworkname{} achieves robust performance, reducing prediction error (MAE) by up to 35.87\%, significantly outperforming state-of-the-art baselines. Our framework bridges the gap between correlational models and causal reasoning, providing a solution that is more accurate and transferable, while also offering crucial interpretability, providing a more reliable foundation for real-world traffic management during critical disruptions.
\end{abstract}


% =============================
% Document
% =============================
\section{Introduction}
\label{sec:intro}

% P1: Hook - Importance & Challenge
Accurate traffic forecasting is fundamental for intelligent urban management, yet its reliability falters during high-impact, non-recurrent events like accidents or extreme weather \cite{ren2022tbsm}. These disruptions cause abrupt deviations from typical patterns, making event-aware forecasting essential for proactive guidance and emergency response \cite{zhang2024traffic}. This high-stakes need demands models that are not only accurate under volatile conditions but also reliable and interpretable for operational decisions. Consequently, advanced data-driven methods, especially deep learning, are under scrutiny to address this challenge.

% P2: SOTA - GNNs
Deep learning approaches, particularly Graph Neural Networks (GNNs), have emerged as the dominant paradigm for traffic prediction. By explicitly modeling complex spatial dependencies via graph structures, GNNs effectively capture traffic propagation and congestion dynamics that traditional sequence models overlook \cite{wang2023st, wu2019graph}.

% P3: The Core Problem - Correlational vs. Causal
The success of GNNs, however, relies on a key assumption: that future patterns will largely resemble historical data. This makes them effective at modeling recurring patterns, such as daily rush hours. Their reliability falters when faced with non-recurring events—disruptions like traffic accidents, road construction, or public concerts \cite{ke2024interpretable}. The core challenge is twofold: these events are often data-sparse, making them difficult to learn from historical patterns, and simultaneously highly heterogeneous in events' type, severity, and duration \cite{chen2025scalable}. This inherent stochasticity means that past disruptions provide an unreliable reference for future occurrences. Furthermore, critical causal evidence is often embedded in unstructured textual reports, which conventional spatiotemporal models cannot parse \cite{ni2016forecasting}.

% P4: Enabling Tech - LLM + Causal
% 1. Rise of LLMs & Capabilities + ST Applications
Recently, Large Language Models (LLMs) have emerged as a powerful paradigm, offering powerful capabilities in semantic understanding, zero-shot reasoning, and knowledge extraction from unstructured text. These attributes have sparked a surge of research in spatiotemporal learning, covering tasks such as forecasting \cite{zhang2024large} and anomaly detection \cite{shao2025towards}. 
% 1. Benefit to Event Prediction (Parsing + Explanation)
For event-driven traffic scenarios, LLMs provide a vital bridge: they can parse complex, unstructured event logs into structured insights and generate natural language explanations, offering indispensable transparency for operational decision-making \cite{guo2024towards}. 
% 2. LLM + Causal Inference
Concurrently, the integration between LLMs and causal inference is gaining attention, showing promise in mining deep causal relationships directly from textual data to uncover the mechanisms behind data variations \cite{liCausalInterventionWhat2025}.

% P5: Gap & Solution
Despite this potential, a critical methodological gap remains in transforming qualitative, LLM-derived causal insights into quantitative, actionable priors for spatiotemporal forecasting. To bridge this gap, we propose \textbf{\frameworkname{}}, a framework integrating LLM-driven feature engineering, formal causal inference, and a novel Causal-Enhanced Prediction Network (CPN). \frameworkname{} is designed to overcome correlational limitations by injecting explicit causal knowledge, enabling robust, accurate, and interpretable forecasts during disruptive events.

% P6: Methodology Breakdown & Validation
Our framework operates in three reinforcing stages: (1) An LLM pipeline converts unstructured event texts into structured causal features; (2) An offline Causal Knowledge Base (CKB) is constructed by estimating the Average Treatment Effects (ATEs) of heterogeneous events; (3) A Causal Injection mechanism dynamically modulates predictions via a novel causal-aware attention layer. This dual-headed architecture decomposes forecasts into a base trend and a measurable causal adjustment. Validated on a real-world multi-modal dataset, \frameworkname{} significantly outperforms SOTA baselines, demonstrating superior long-term robustness and verifiable, causal-driven interpretability.

% P7: Contributions
The main contributions of this work are:
\begin{itemize}
    \item We propose a novel pipeline that integrates LLMs and formal causal inference to construct a quantitative CKB from unstructured text.
    \item We design a CPN with a causal-aware attention mechanism that explicitly decomposes correlational trends and causal effects.
    \item We provide extensive validation on a real-world dataset, demonstrating that \frameworkname{} achieves SOTA accuracy, long-term robustness, and verifiable interpretability.
\end{itemize}

% P8: Roadmap
The rest of this paper is organized as follows. Section~\ref{sec:methodology} details the theoretical foundations and design of \frameworkname{}. Section~\ref{sec:exp} presents the experimental setup, comparative analysis, and in-depth discussions on causal effects and interpretability. Section~\ref{sec:conclusion} concludes the paper and outlines future directions.

\section{Methodology}
\label{sec:methodology}
This section details our proposed \frameworkname{}, a framework integrating LLMs for event understanding, formal causal inference for knowledge extraction, and a novel graph-attentional network for robust, event-aware traffic forecasting. Our approach is structured around two core phases: an Offline Causal Knowledge Base Construction and an Online Causal-Enhanced Prediction Network.

\subsection{Problem Formulation}

We formulate the traffic speed prediction problem as a segment-wise, graph-augmented time-series forecasting task. We define the road network as a directed graph $G=(V,E,A)$, where $V$ is the set of road segments (nodes), $E$ represents connectivity, and $A \in \mathbb{R}^{|V| \times |V|}$ is the adjacency matrix.

Given a historical lookback window of length $T$, the inputs for predicting the target speed $y_i$ for segment $i$ are:

\textit{Target Segment Features}: The feature sequence of the target segment $i$, $\mathbf{x}_i = (x_{i,t-T+1}, ..., x_{i,t}) \in \mathbb{R}^{T \times D_t}$, which includes historical speed and temporal encodings.

\textit{Dynamic Causal Features}: An event-based feature sequence $\mathbf{e}_i = (e_{i,t-T+1}, ..., e_{i,t}) \in \mathbb{R}^{T \times D_c}$, derived from textual event logs $E_{\text{text}}$ and our CKB.

\textit{Neighborhood Features}: The feature sequences of the Top-K neighbors $\mathcal{N}_K(i)$ of segment $i$, $\{\mathbf{x}_j\}_{j \in \mathcal{N}_K(i)}$, and their corresponding adjacency weights $\{w_{ij}\}_{j \in \mathcal{N}_K(i)}$.

Therefore, our goal is to learn a mapping function $f_\theta$ that predicts the future average speed $\hat{y}_i$ for the target segment $i$:
\begin{equation}
f_\theta: (\mathbf{x}_i, \mathbf{e}_i, \{\mathbf{x}_j\}_{j \in \mathcal{N}_K(i)}, \{w_{ij}\}_{j \in \mathcal{N}_K(i)}) \rightarrow \hat{y}_i
\label{eq:problem_formulation}
\end{equation}
The core challenge lies in transforming unstructured $E_{\text{text}}$ into a quantitative causal feature $\mathbf{e}_i$ and integrating it effectively with the spatiotemporal features.

\subsection{Framework Overview}

\begin{figure}[h!]
\centering
\includegraphics[width=1\linewidth]{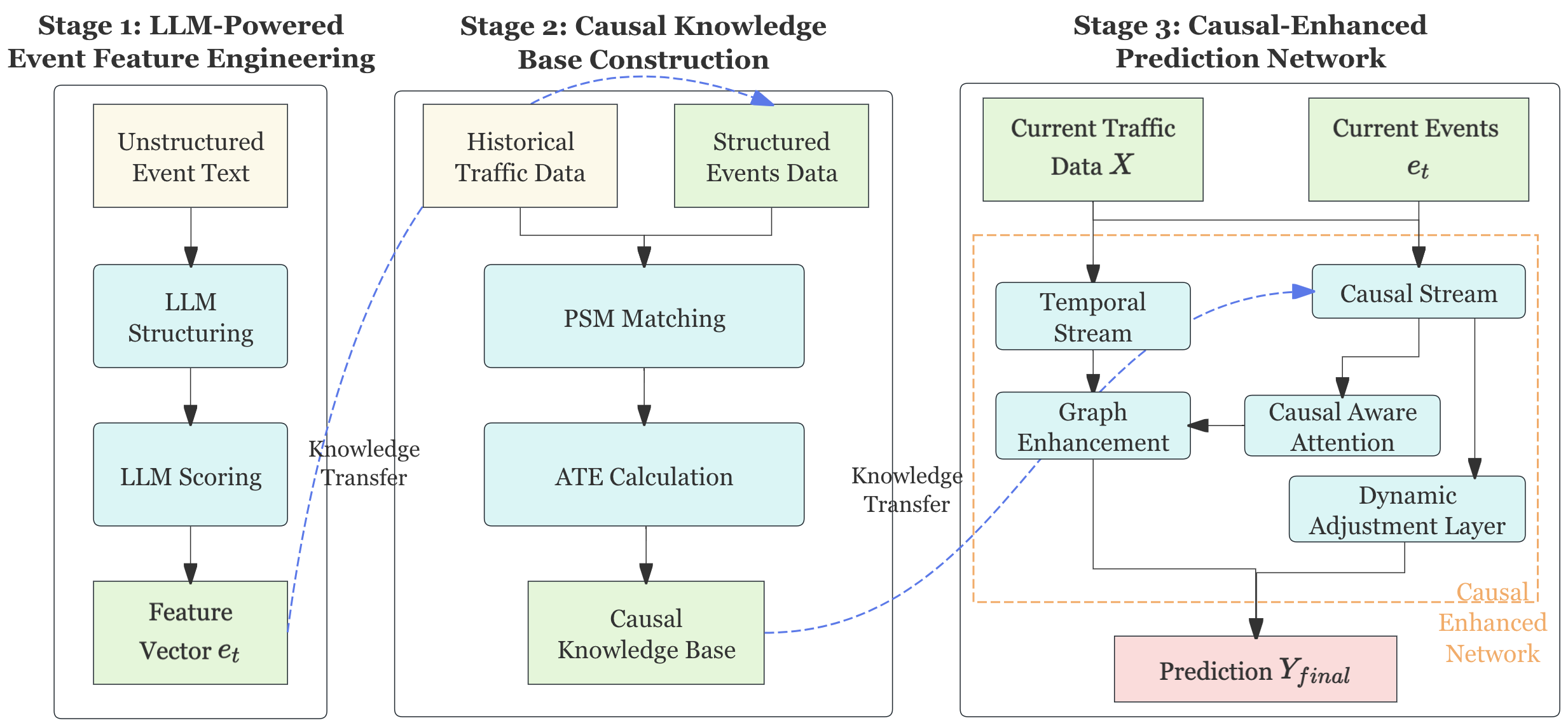}
\caption{The overall architecture of our \frameworkname{}.}
\label{fig:framework}
\end{figure}

As illustrated in~\ref{fig:framework}, \frameworkname{} operates via a multi-stage pipeline consisting of three main components:
\begin{enumerate}
\item LLM-Powered Event Feature Engineering: An offline pipeline where raw event texts $E_{\text{text}}$ are processed by a two-stage LLM to be structured and quantified.
\item Offline CKB Construction: Using the structured event data, we employ Propensity Score Matching (PSM) \cite{caliendo2008some} to estimate the heterogeneous ATE of each event type. These results are stored in the CKB as a general causal prior.
\item Online Causal-Enhanced Prediction: This real-time model, trained on target segments, queries the CKB to fetch the relevant causal prior ($\tau$). This prior is dynamically adjusted by real-time factors (e.g., severity, time decay) to create the causal feature vector $\mathbf{e}_i$. This vector, alongside spatiotemporal features, is fed into the CPN to produce the final, causally-informed forecast.
\end{enumerate}

\subsection{LLM-Powered Event Feature Engineering}
We leverage a two-stage LLM prompting strategy to transform raw, non-standardized traffic event texts into machine-readable, quantified features. This serves as the input for both offline CKB construction and online prediction.

\textit{Stage 1: Event Structurization}: The first LLM prompt parses raw event descriptions to extract key attributes into a structured format, including event time, event type (e.g., Accident, Construction), and location identifiers.

\textit{Stage 2: Event Quantification}: The second LLM prompt takes the structured event and, using a detailed rubric, assigns integer scores (1-5 scale) for severity, danger, duration, and impact scope, while also estimating capacity reduction. These quantitative metrics serve as crucial confounders for causal inference and as inputs for dynamic feature adjustment.

\subsection{Offline Causal Knowledge Base Construction}
\label{sec:ckb_construction}
This offline phase builds a robust, queryable database of quantitative causal impacts, which serves as a strong prior for the online model.

To robustly estimate the causal effect of various event types (treatments) on traffic outcomes (speed change), we leverage a large dataset from distinct road segments (not used in the final prediction task). To isolate the ATE of an event while controlling for confounding factors, we employ PSM. PSM is a statistical method designed to mitigate selection bias in observational data by creating a matched sample of treated units (e.g., segments with an accident) and control units (segments without an accident) that share similar observable characteristics, thereby simulating a randomized controlled trial \cite{caliendo2008some}.

Our implementation of PSM involves two primary stages. First, for a given event type $T$ (e.g., $T=1$ for Accident) and a set of confounders $\mathbf{X}$ (e.g., time period, severity, duration), we estimate the propensity score $e(\mathbf{X})$—the probability of receiving the treatment given the confounders—using logistic regression:
\begin{equation}
e(\mathbf{X}) = P(T=1 | \mathbf{X}) = \frac{1}{1 + e^{-(\beta_0 + \beta^T \mathbf{X})}}
\label{eq:propensity_score}
\end{equation}

Second, treated samples are then matched with control samples based on their propensity scores, using 1:1 nearest-neighbor matching within a predefined caliper. The ATE is then estimated from the matched pairs $M$:
\begin{equation}
\hat{\text{ATE}} = \frac{1}{|M|} \sum_{(i,j) \in M} (Y_{i}^{\text{treated}} - Y_{j}^{\text{control}})
\label{eq:ate}
\end{equation}

We compute this ATE for heterogeneous groups (e.g., event type $\times$ time period). The resulting estimates (e.g., $\hat{\text{ATE}}_{\text{Speed, Accident, MorningPeak}} = -15.2 \text{ km/h}$) are stored in the CKB. This CKB, reflecting population-level average effects, is then used as a fixed prior for training the prediction model on a separate, unseen set of target segments.
\subsection{Causal-Enhanced Prediction Network}
\label{sec:cpn}
The CPN serves as the online inference engine. It is trained end-to-end to integrate real-time traffic observations, spatial neighborhood information, and dynamic causal features derived from the CKB.

\subsubsection{Dynamic Causal Feature Engineering}
\label{sec:causal_feature_engineering}
During online inference, when a new event $i$ occurs at time $t$, the CPN engineers a dynamic causal feature vector $\mathbf{e}_{i,t} \in \mathbb{R}^{D_c}$, which serves as the primary causal input to the network. This process involves several steps:

\textit{Step1: CKB Query.} We retrieve the base causal effect vector $\boldsymbol{\tau}_i$ from the CKB based on the event's type and the current time period.

\textit{Step2: Dynamic Adjustment.} This base ATE is modulated by real-time event attributes (from the LLM) and temporal decay. For each time step $t$, we sum the influence of all active events:
    \begin{equation}
    \mathbf{e}_{i,t}^{\text{adjust}} = \sum_{j \in \text{ActiveEvents}} w_j^{\text{temporal}} \cdot w_j^{\text{severity}} \cdot w_j^{\text{danger}} \cdot \boldsymbol{\tau}_j
    \label{eq:causal_feature}
    \end{equation}
    where $w_j^{\text{temporal}} = \exp(-\Delta t_j / \lambda)$ models temporal decay, and $w^{\text{severity}}, w^{\text{danger}}$ are weights derived from the LLM's quantification. The final feature $\mathbf{e}_{i,t}$ is then composed of this adjustment $\mathbf{e}_{i,t}^{\text{adjust}}$, event counts, time since last event, and effect confidence.
\subsubsection{CPN Architecture}
\label{sec:cpn_architecture}

The CPN is a multi-component network designed to fuse temporal, spatial, and causal information.

\textbf{a. Temporal Encoder.}
For each road $i$, we employ a Gated Recurrent Unit (GRU) to capture the temporal evolution of its traffic features $\mathbf{x}_{i,t}$ over the lookback window $T$:
\begin{equation}
\mathbf{h}_{i,t} = \text{GRU}(\mathbf{x}_{i,t}, \mathbf{h}_{i,t-1})
\label{eq:gru}
\end{equation}
where $\mathbf{h}_{i,t} \in \mathbb{R}^{d_h}$ is the temporal hidden state. A sinusoidal position encoding is added to $\mathbf{h}_{i,t}$ to create the final temporal representation $\mathbf{z}_{i,t}$.

\textbf{b. Lightweight Graph-Neighbor Encoder.}
To incorporate spatial dependencies with linear complexity, we use a lightweight, sampling-based neighbor encoder.

\textit{Neighbor Sampling}: For target segment $i$, we sample its Top-5 neighbors $\mathcal{N}_K(i)$ based on the adjacency matrix $A$.

\textit{Neighbor Encoding}: We encode each neighbor's feature sequence $\mathbf{x}_{j,t}$ using the same GRU encoder to get $\mathbf{h}_{j,t}$. Critically, gradients are detached during this step to reduce computational load.

\textit{Graph-Prior Attention}: A lightweight graph attention layer aggregates these neighbor representations $\mathbf{h}_{j,t}$ into a single spatial context vector $\mathbf{g}_{i,t}$. The attention score $a_{ij}$ is explicitly biased by the normalized adjacency weight $w_{ij}$ from $A$, injecting the graph structure as a strong prior:
\begin{equation*}
e_{ij} = \text{AttentionScore}(\mathbf{z}_{i,t}, \mathbf{h}_{j,t}) + \log(w_{ij})
\end{equation*}

\textit{Spatiotemporal Fusion}: The aggregated graph context $\mathbf{g}_{i,t}$ is then fused with the target segment's temporal representation $\mathbf{z}_{i,t}$ using a final attention layer, yielding the spatiotemporal representation $\mathbf{z}_{i,t}^{\text{st}} \in \mathbb{R}^{d_h}$.

\textbf{c. Causal-Aware Attention Mechanism.}
We design a causal-aware attention mechanism that uses the causal feature stream $\mathbf{e}_{i,t}$ to modulate the spatiotemporal stream $\mathbf{z}_{i,t}^{\text{st}}$. The causal feature sequence $\mathbf{E}_i = \{\mathbf{e}_{i,1}, \ldots, \mathbf{e}_{i,T}\}$ is first projected by an MLP: $\tilde{\mathbf{c}}_{i,t} = \text{MLP}(\mathbf{e}_{i,t})$. This causal signal $\tilde{\mathbf{c}}$ is injected in two ways:

\textit{Channel Fusion:} The causal embedding is additively fused with the spatiotemporal representation: $\mathbf{z}_{i,t}^{\text{fused}} = \text{LayerNorm}(\mathbf{z}_{i,t}^{\text{st}} + \tilde{\mathbf{c}}_{i,t})$. This combined representation is then used to generate the attention queries ($Q$), keys ($K$), and values ($V$).

\textit{Attention Score Biasing:} We inject a learnable causal bias into the attention score calculation. Let $m_s \in \{0, 1\}$ be a mask (1 if an event exists at time $s$). We modify the standard attention score $e_{t,s}$ with an autoregressive mask and this event bias:
\begin{equation}
e'_{t,s} = \frac{\mathbf{q}_t \cdot \mathbf{k}_s}{\sqrt{d_k}} + \beta \cdot m_s + \mathbf{M}_{\text{auto}}
\label{eq:attention_alpha}
\end{equation}
where $\beta$ is a learnable scalar, and $\mathbf{M}_{\text{auto}}$ is the autoregressive mask (setting future steps to $-\infty$). This forces the model to pay more attention to timestamps where critical events occurred.

The output of this module is the causally-aware hidden state $\tilde{\mathbf{z}}_T$ from the final time step.

\textbf{d. Causal Adjustment Prediction Head.}
We use a dual-headed architecture to explicitly decompose the prediction. The final state $\tilde{\mathbf{z}}_T$ is fed into two separate MLPs:

\textit{Base Predictor} ($f_{base}$): Predicts the base trend $y_{base}$, representing the speed if no event had occurred.

\textit{Causal Estimator} ($f_{causal}$): Predicts the causal adjustment $a_{causal}$, representing the deviation from the base trend caused by the event.

\subsubsection{Progressive Training Strategy}
\label{sec:training_strategy}
To stabilize training, we adopt a three-phase progressive strategy. Initially, the model learns fundamental spatiotemporal patterns by training only the temporal and graph modules on $\mathcal{L}_{\text{MSE}}$, while all causal components are frozen. Subsequently, the causal modules are unfrozen and introduced via a weakly-gated fusion, allowing the model to gently learn the causal adjustment ($a_{causal}$) using the full loss function. In the final phase, the fusion gate is fully activated, enabling the model to fine-tune the optimal balance between the base trend and the causal adjustment.

\subsubsection{Loss Function}
\label{sec:loss}
The CPN is optimized using a composite loss function designed to balance prediction accuracy with causal plausibility, defined as:
\begin{equation}
\mathcal{L} = \mathcal{L}_{\text{MSE}} + \beta \cdot \mathcal{L}_{\text{causal}} + \gamma \cdot \mathcal{L}_{\text{entropy}}
\label{eq:composite_loss}
\end{equation}
The three components are:

\textit{Prediction Loss ($\mathcal{L}_{\text{MSE}}$).}
The primary objective is the standard MSE between the final predicted speed $\hat{y}$ and the ground truth $y$:
\begin{equation}
\mathcal{L}_{\text{MSE}} = | \hat{y} - y |^2
\label{eq:loss_mse}
\end{equation}

\textit{Causal Consistency Loss ($\mathcal{L}_{\text{causal}}$).}
This is the core regularizer, designed to enforce causal plausibility by ensuring the model's explicit causal explanation aligns with the observed outcome. Specifically, $\mathcal{L}_{\text{causal}}$ forces the sign (direction) of the model's internal causal adjustment ($a_{\text{causal}}$) to match the sign of the self-supervised residual ($y - y_{\text{base}}$). This ensures the model is right for the reasons, as its explicit adjustment must correspond to the observed traffic change caused by the event.
\begin{equation}
\mathcal{L}_{\text{causal}} = \mathbb{E}_{i \in \text{Events}} \left[ \mathcal{L}_{\text{sign}}(\text{sign}(a_{\text{causal}}^{(i)}), \text{sign}(y^{(i)} - y_{\text{base}}^{(i)})) \right]
\label{eq:loss_causal}
\end{equation}
Here, $y^{(i)}$ is the ground truth speed, $y_{\text{base}}^{(i)}$ is the model's own "what if" prediction generated without causal features, and $\mathcal{L}_{\text{sign}}$ is the sign-matching loss (e.g., MSE or Cross-Entropy).

\textit{Attention Entropy Loss ($\mathcal{L}_{\text{entropy}}$).}
To prevent the causal-aware attention distribution $\alpha$ from collapsing to a single time step and encourage a diverse policy, we apply an entropy regularization term:
\begin{equation}
\mathcal{L}_{\text{entropy}} = \sum_{t=1}^{T} \sum_{s=1}^{T} \alpha_{t,s} \log(\alpha_{t,s})
\label{eq:loss_entropy}
\end{equation}
where $\alpha_{t,s}$ represents the attention weight from time step $s$ in the lookback window to prediction step $t$.

\section{Experiments}
\label{sec:exp}

\subsection{Datasets}
\label{sec:dataset}
We use the \textit{Beijing Text-Traffic (BjTT)} dataset \cite{zhang2024bjtt}, a real-world, multi-modal collection of two temporally-aligned sources from Beijing:

    \textit{Spatio-Temporal Traffic Data}: A one-month subset of high-frequency (4-minute interval) traffic velocity recordings from all $N=1260$ major road segments (vertices).
    
    \textit{Unstructured Event Logs}: 4,215 corresponding textual event reports (e.g., accidents, construction) sourced from social media and map applications.

The dataset precisely aligns each 4-minute traffic record with its corresponding textual event report, forming the input for our framework.

\subsection{Experiment Settings}
\label{sec:settings}

\subsubsection{Implementation and Evaluation}
Experiments are implemented in PyTorch 2.1.2 on an NVIDIA RTX 4090D GPU. We use a chronological 70:15:15 split for training, validation, and testing. We report four standard metrics (MAE, RMSE, MAPE) calculated on the original data scale. We compare our CPN against SOTA baselines.

\subsubsection{Hyperparameter Details}
Our model uses a 1-hour lookback window ($T=15$ steps) to predict the next 12, 24, 32 minutes ($H=3,6,8$ steps). We sample $K=5$ neighbors for the graph module. The CPN architecture features a hidden dimension of $d=64$, $L=2$ layers, and $N_{heads}=4$ attention heads. We train for 1000 epochs (batch size 256) using the Adam optimizer (LR $3 \times 10^{-5}$, Weight Decay $1 \times 10^{-2}$) and our Progressive Training Strategy. Early stopping (patience=50) and gradient clipping (norm=5.0) are applied.
% (NOTE: The detailed loss weights (alpha, beta, gamma), scheduler details, and warmup strategies are excellent candidates for an Appendix to streamline the main paper.)
% \textit{Further implementation details, including loss weights and scheduler settings, are provided in the Appendix.}

\subsection{Baseline Methods}
\label{sec:baselines}
We compare CPN against several SOTA spatiotemporal GNNs:
\begin{itemize}
    \item \textbf{STGCN}~\cite{yu2017spatio}: Integrates GCNs with 1D temporal convolutions.
    \item \textbf{DCRNN}~\cite{li2018dcrnn}: Utilizes diffusion graph convolutions within an encoder-decoder framework.
    \item \textbf{ASTGCN}~\cite{guo2019attention}: Employs spatial and temporal attention mechanisms for dynamic weighting.
    \item \textbf{GraphWaveNet}~\cite{wu2019graph}: Combines graph convolutions with causal convolutions and a self-adaptive adjacency matrix.
\end{itemize}

\subsection{Causal Knowledge Base Validation}
\label{sec:causal_results}

We first validate the CKB by quantifying the ATE of traffic events on traffic speed using PSM, controlling for key confounders (e.g., severity, duration, time period). 
Table~\ref{tab:causal_effects} summarizes these findings, which are stored in the CKB as causal priors. The results, derived from a large number of matched samples, are statistically significant and stable.

% --- Causal Effect Table (ATE, SE, and Significance) ---
\begin{table*}[h!]
\centering
\caption{Causal Effect (ATE) of Traffic Events on Vehicle Speed (km/h), by Time of Day. Standard Errors are in parentheses.}
\label{tab:causal_effects}
\scalebox{0.95}{ %
\small
\begin{tabular}{l|cccc}
\toprule
\textbf{Event Type} & \textbf{Morning Peak} & \textbf{Evening Peak} & \textbf{Off-peak} & \textbf{Night} \\
\midrule
Accident & 5.70*** (0.05) & 6.58*** (0.07) & -1.50*** (0.03) & 1.29*** (0.11) \\
Construction & 6.48*** (0.21) & -5.23*** (0.14) & -4.63*** (0.08) & -3.07*** (0.12) \\
Hazard & -10.06*** (0.51) & 0.89** (0.32) & -2.94*** (0.31) & -4.20*** (0.52) \\
Road Closure & -2.42*** (0.19) & -1.52*** (0.32) & 3.88*** (0.11) & 18.32*** (0.20) \\
Traffic Control & -2.09*** (0.15) & -1.43*** (0.11) & -1.51*** (0.05) & -1.18*** (0.09) \\
\bottomrule
\multicolumn{5}{l}{Significance levels: *** $p < 0.001$, ** $p < 0.01$, * $p < 0.05$} \\
\end{tabular}
}
\end{table*}
% ----------------------------------------

The analysis reveals several key patterns. Some events, such as Traffic Control (ATEs from -1.18 to -2.09) and Construction, exhibit consistent negative effects on speed across all time periods. In terms of magnitude, the most severe negative impact observed is from Hazard events during the Morning Peak (ATE: -10.06), while the strongest positive impact comes from Road Closures at Night (ATE: +18.32). The analysis also highlights significant heterogeneous effects: the impact of Accidents and Road Closures is highly time-dependent, with their ATEs shifting direction (positive vs. negative) based on the time of day. While the positive ATE for peak-hour accidents (+5.70) appears counter-intuitive, it is consistent with the effects of proactive navigation re-routing, which diverts traffic, resulting in higher speeds on the affected segment compared to its typically congested control counterparts

\begin{figure}[h!]
  \centering
  \includegraphics[width=0.8\linewidth]{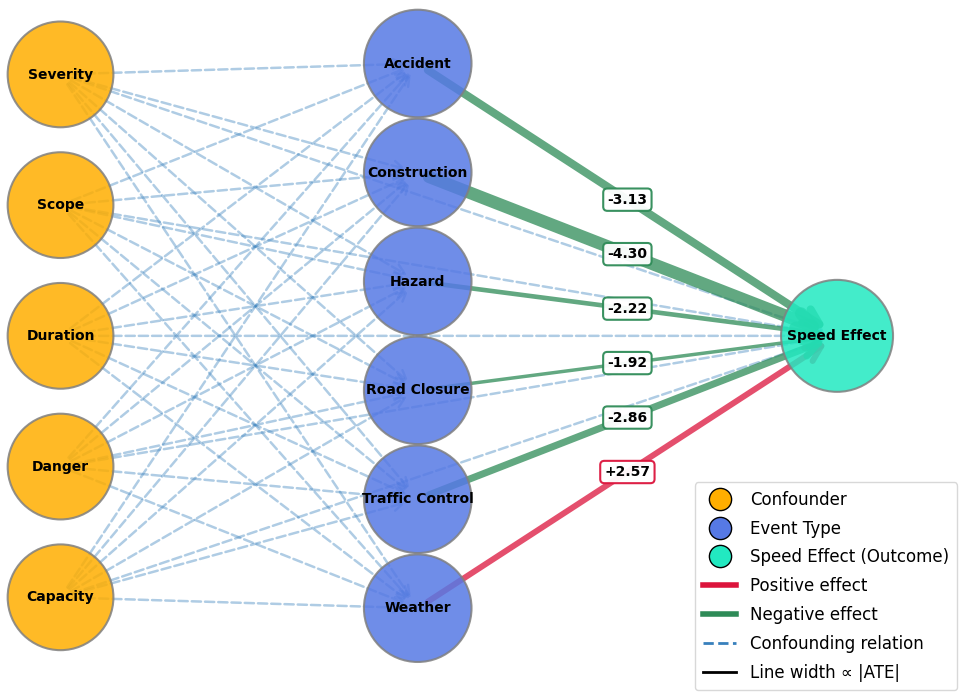}
  \caption{The Causal DAG of Total Effects derived from the CKB.}
  \label{fig:dag}
\end{figure}

\subsection{Overall Performance}
\label{sec:overall_performance}

We evaluate CPN against baselines for short- (H=3, 12 min), medium- (H=6, 24 min), and long-term (H=8, 32 min) forecasting, with a fixed lookback of $T=15$ (60 min). 

Table~\ref{tab:overall_performance} shows that CPN significantly outperforms all SOTA baselines across all horizons and metrics. For short-term prediction (H=3), CPN (MAE 3.607) already achieves a 10.0\% improvement over the strongest baseline, GraphWaveNet (MAE 4.008).

% --- Overall Performance Table ---
\begin{table*}[h!]
\centering
\caption{Overall Prediction Performance Comparison on Traffic Speed Forecasting. The input sequence length is $T=15$. 
\textbf{Bold} indicates the best result, and \underline{underline} indicates the second-best result.}
\label{tab:overall_performance}
\scalebox{0.92}{%
\small
\begin{tabular}{l|ccc|ccc|ccc}
\toprule
\multirow{2}{*}{\textbf{Model}} & \multicolumn{3}{c|}{$\mathbf{H=3}$ (12 min)} & \multicolumn{3}{c|}{$\mathbf{H=6}$ (24 min)} & \multicolumn{3}{c}{$\mathbf{H=8}$ (32 min)} \\
\cline{2-10}
& MAE & MSE & RMSE & MAE & MSE & RMSE & MAE & MSE & RMSE \\
\midrule
STGCN & 4.013 & 35.239 & 5.936 & \underline{3.979} & \underline{33.140} & \underline{5.757} & 3.983 & 33.315 & 5.772 \\
ASTGCN & 4.041 & 37.351 & 6.112 & 4.288 & 44.152 & 6.645 & 4.410 & 48.205 & 6.943 \\
DCRNN & 4.022 & 37.399 & 6.116 & 4.575 & 49.044 & 7.003 & 4.622 & 49.850 & 7.060 \\
GraphWaveNet & \underline{4.008} & \underline{37.412} & \underline{6.117} & 4.091 & 38.516 & 6.206 & \underline{4.200} & \underline{40.392} & \underline{6.355} \\
\midrule
\textbf{CPN (Ours)} & \textbf{3.607} & \textbf{25.427} & \textbf{5.043} & \textbf{3.113} & \textbf{21.801} & \textbf{4.669} & \textbf{2.964} & \textbf{20.657} & \textbf{4.545} \\
\bottomrule
\end{tabular}
}
\end{table*}

% ----------------------------------------

This superiority widens at longer horizons. While the performance of all SOTA baselines degrades as the prediction horizon increases with longer-term forecasting, CPN exhibits strong robustness. 

Notably, CPN's prediction error decreases at longer horizons, with its MAE improving to 2.964 at $H=8$ (32 minutes). This widens the performance gap substantially: at $H=8$, CPN's MAE is 25.6\% lower than the best-performing baseline, STGCN (MAE 3.983). 
This demonstrates that the integration of causal-enhanced attention and causal adjustment effectively leverages event-driven features, which provide stable, high-quality predictive signals crucial for accurate long-term forecasting.

\begin{figure}[h!]
    \centering
    \begin{subfigure}[b]{0.45\columnwidth}
    \includegraphics[width=\linewidth]{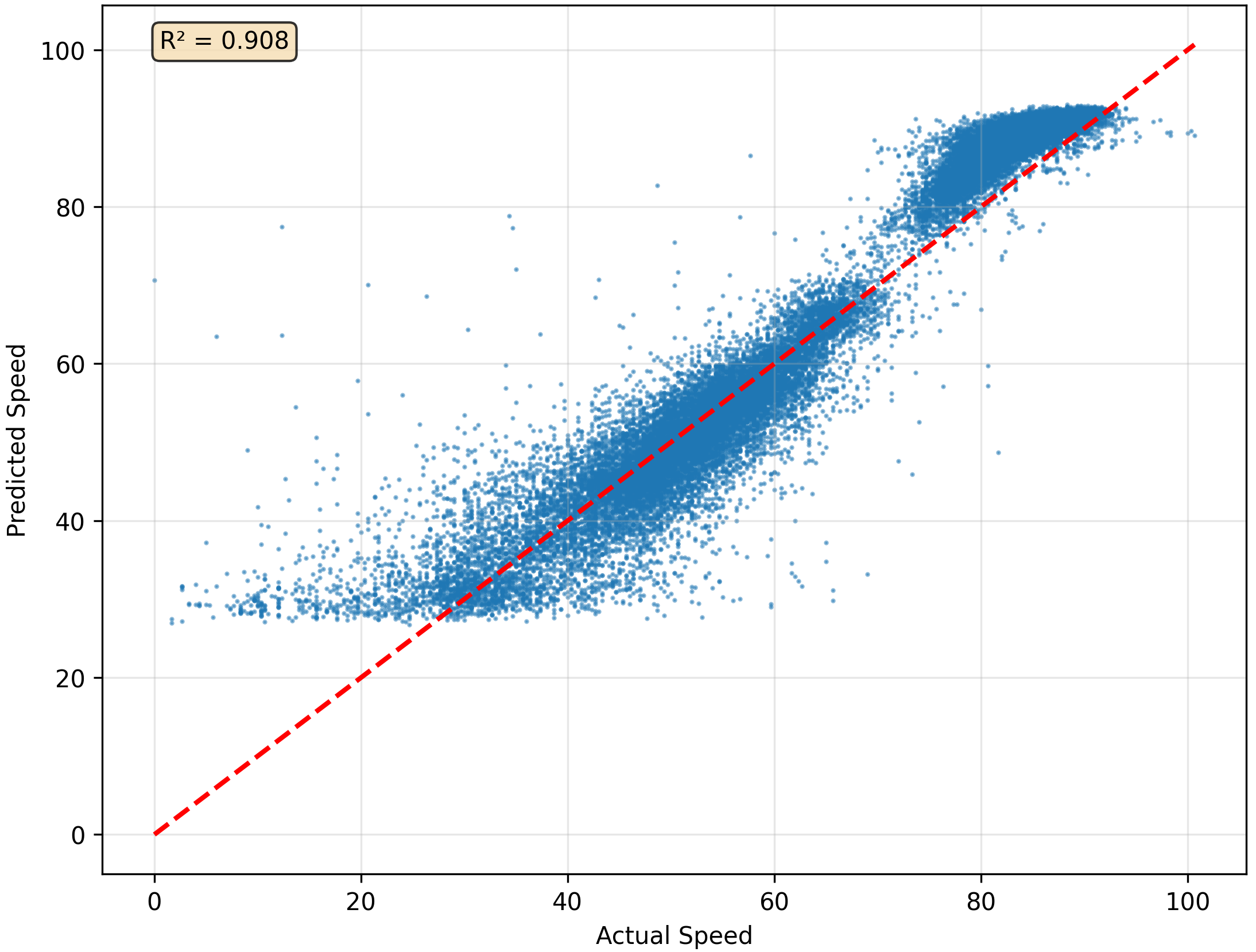}
        \caption{Prediction vs. Actual}
        \label{fig:diag_scatter}
    \end{subfigure}
   % \hfill 
    \begin{subfigure}[b]{0.45\columnwidth}
    \includegraphics[width=\linewidth]{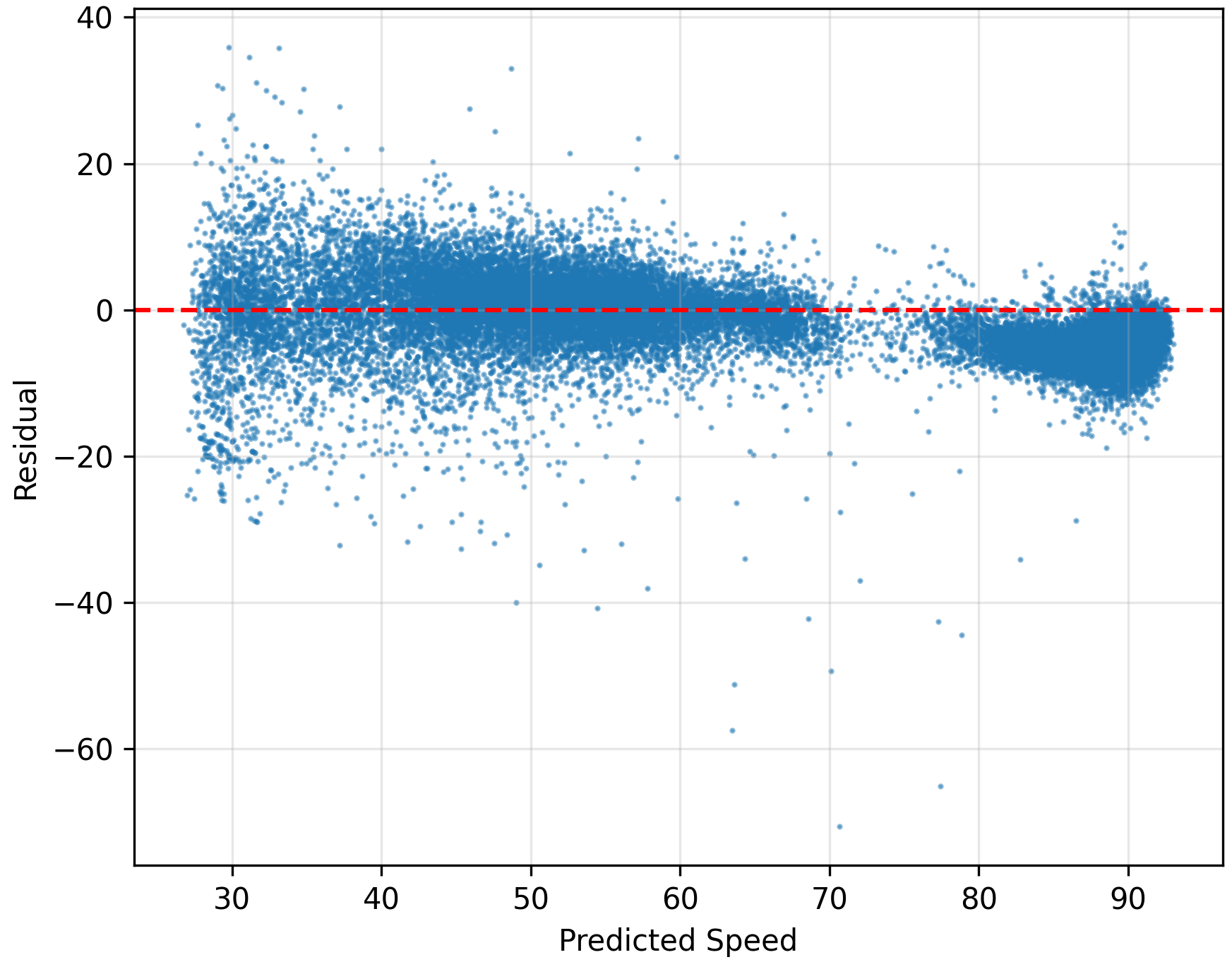} 
            \caption{Residuals vs. Predictions}
            \label{fig:diag_residual}
    \end{subfigure}
    \caption{Diagnostic plots for CPN.}
    \label{fig:diagnostic_plots} % 
\end{figure}

Further diagnostic plots in Figure~\ref{fig:diagnostic_plots} confirm this robustness. The scatter plot (Fig.~\ref{fig:diagnostic_plots}(a)) shows high accuracy, with predictions tightly clustered on the 45-degree ideal line ($\mathbf{R^2}=0.908$). Crucially, the residual analysis (Fig.~\ref{fig:diagnostic_plots}(b)) demonstrates model stability: the residuals exhibit clear homoscedasticity, forming a tight, random cloud centered at zero with no systemic correlation to predicted speed. This visually confirms CPN's error is low, stable, and unbiased, proving its reliability during critical, event-driven scenarios.

\subsection{Model Interpretability}
\label{sec:interpretability}

We analyze CPN's internal attention mechanism to verify its interpretability and understand how causal integration shapes its policy.

% --- Figure Caption (Streamlined) ---
\begin{figure}[h!]
    \centering
    % (a) Subfigure for Average Attention
    \begin{subfigure}[b]{0.46\columnwidth}
        \includegraphics[width=\linewidth]{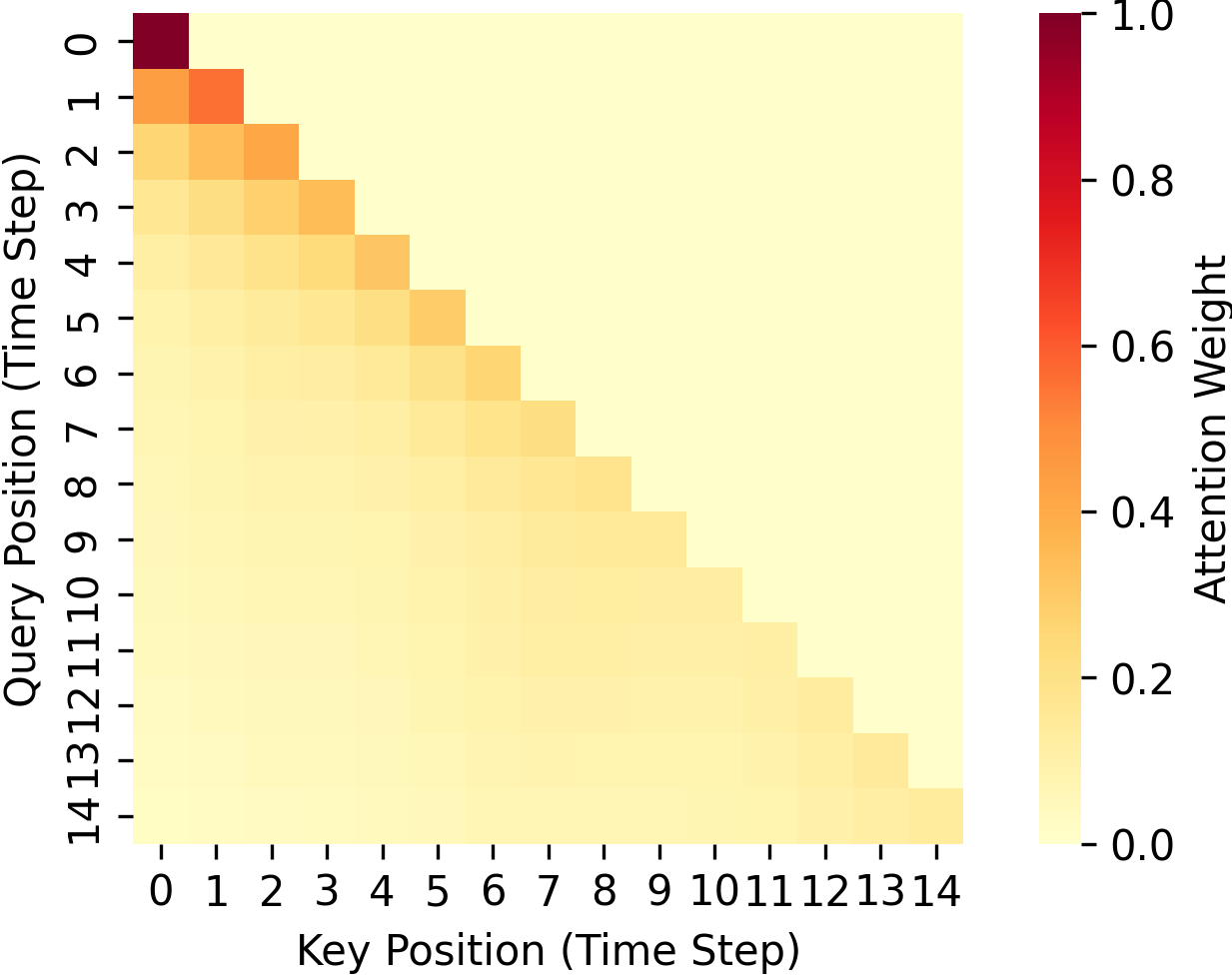}
        \caption{Average attention map}
        \label{fig:attn_avg}
    \end{subfigure}
    \hfill % 
    % (b) Subfigure for Attention Variance
    \begin{subfigure}[b]{0.49\columnwidth}
        \includegraphics[width=\linewidth]{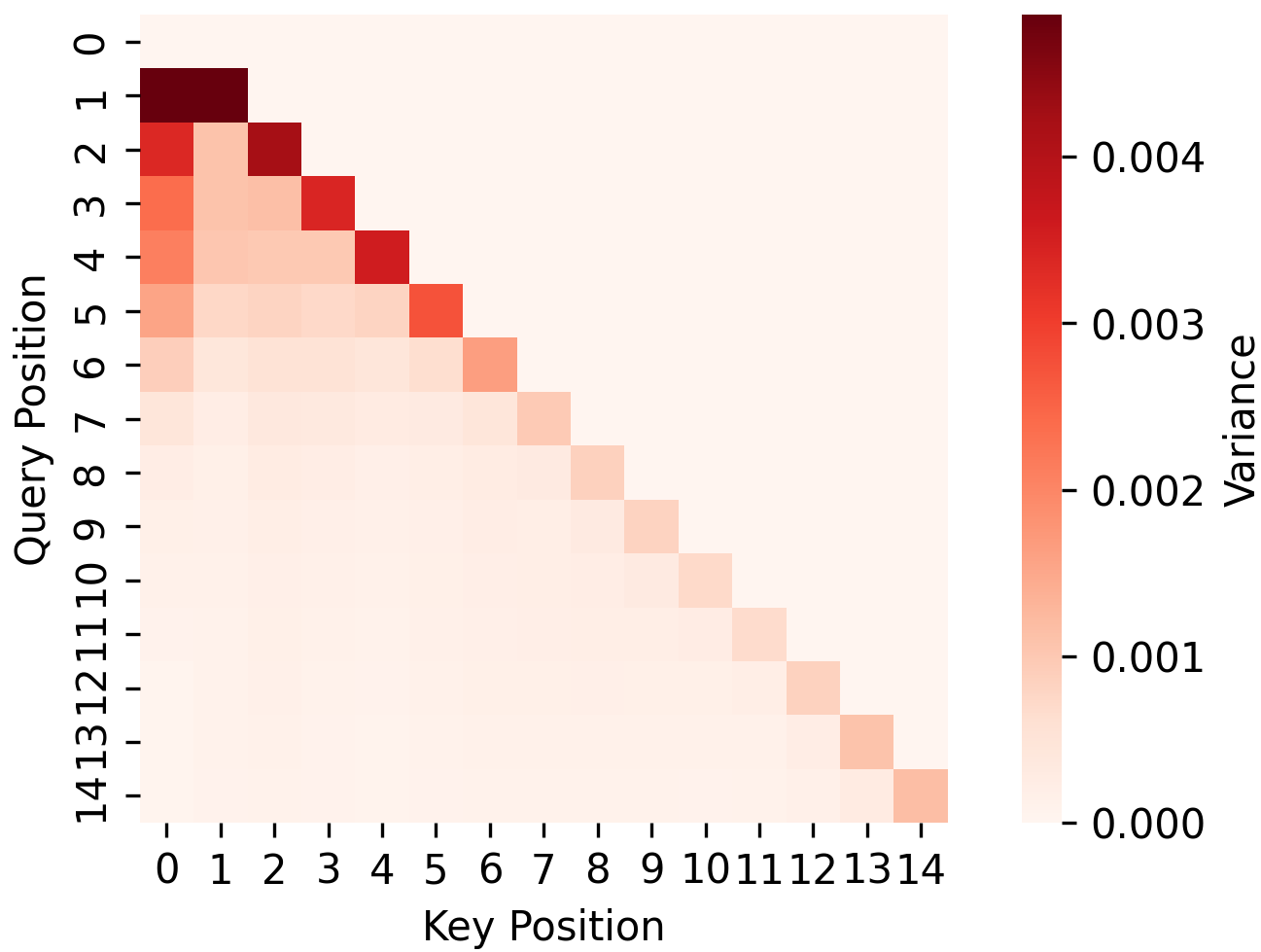}
        \caption{Attention variance map}
        \label{fig:attn_var}
    \end{subfigure}
    
    \caption{CPN's learned attention dynamics.}
    \label{fig:attn} %
\end{figure}

% -----------------------------------

Figure~\ref{fig:attn}(a) reveals CPN's highly localized attention strategy, with weights concentrated overwhelmingly on the current time step ($t=0$). This indicates the model learns to \textit{trust} the current, causally-informed feature vector, which renders a noisy search of historical data unnecessary.

The attention variance (Fig.~\ref{fig:attn}(b)) further reveals a key policy dichotomy. Near-zero variance (bright white) in off-diagonal regions confirms a stable policy of ignoring the past, enabled by the causal context. Conversely, high variance (dark red) concentrated on the main diagonal demonstrates strategic flexibility, allowing the model to dynamically adjust its focus based on the current event's immediate importance.

This duality—structural stability (consistently ignoring the noisy past) and strategic flexibility (adaptively responding to the present event)—directly explains CPN's superior long-term forecasting robustness.

\section{Conclusions and Future Work}
\label{sec:conclusion}

In conclusion, we introduce \frameworkname{}, a novel framework for robust and interpretable traffic prediction, especially during non-recurring disruptions. By bridging correlational models with causal reasoning, \frameworkname{} achieves superior performance over SOTA baselines while providing crucial interpretability. Its dual-stream architecture, leveraging an LLM-built causal knowledge base and a causal attention mechanism, provides a transparent and robust solution for critical traffic management, forming a reliable foundation for decision-making during urban disruptions.

Future work will focus on two key areas. First, we aim to investigate the transferability and generalization of the CKB across different cities and network topologies, which is essential for adapting causal estimates to new environments. We also plan to incorporate a broader set of external causal factors—such as extreme weather, public events, and policy changes—to enhance forecast comprehensiveness and accuracy. Second, a significant long-term direction is the development of large-scale causal reasoning systems for urban brains. This vision entails integrating city-level, multi-modal causal data in real-time, and will require addressing substantial challenges in computational resources and engineering reliability to support city-wide traffic planning and management.

% \section{Acknowledgment}
% This work is supported by the Kunshan Municipal Special Project under Grant No. 24KKSGR024, the Professional Discretionary Fund 26AKUG0088, and the Mitacs Elevate Canada under Grant IT44479.
% --- References ---
% =============================
\bibliographystyle{IEEEtran}
\bibliography{rfrs}

@article{ren2022tbsm,
  title={TBSM: A traffic burst-sensitive model for short-term prediction under special events},
  author={Ren, Yilong and Jiang, Han and Ji, Nan and Yu, Haiyang},
  journal={Knowledge-Based Systems},
  volume={240},
  pages={108120},
  year={2022},
  publisher={Elsevier}
}

@article{zhang2024traffic,
  title={A traffic-weather generative adversarial network for traffic flow prediction for road networks under bad weather},
  author={Zhang, Wensong and Yao, Ronghan and Yuan, Ying and Du, Xiaojing and Wang, Libing and Sun, Feng},
  journal={Engineering Applications of Artificial Intelligence},
  volume={137},
  pages={109125},
  year={2024},
  publisher={Elsevier}
}

@article{zhang2024bjtt,
  title={BjTT: A large-scale multimodal dataset for traffic prediction},
  author={Zhang, Chengyang and Zhang, Yong and Shao, Qitan and Feng, Jiangtao and Li, Bo and Lv, Yisheng and Piao, Xinglin and Yin, Baocai},
  journal={IEEE Transactions on Intelligent Transportation Systems},
  year={2024},
  publisher={IEEE}
}

@article{yu2017spatio,
  title={Spatio-temporal graph convolutional networks: A deep learning framework for traffic forecasting},
  author={Yu, Bing and Yin, Haoteng and Zhu, Zhanxing},
  journal={arXiv preprint arXiv:1709.04875},
  year={2017}
}

@inproceedings{wang2023st,
  title={ST-GIN: An uncertainty quantification approach in traffic data imputation with spatio-temporal graph attention and bidirectional recurrent united neural networks},
  author={Wang, Zepu and Zhuang, Dingyi and Li, Yankai and Zhao, Jinhua and Sun, Peng and Wang, Shenhao and Hu, Yulin},
  booktitle={2023 IEEE 26th international conference on intelligent transportation systems (ITSC)},
  pages={1454--1459},
  year={2023},
  organization={IEEE}
}

@article{zhang2024large,
  title={Large language models for mobility analysis in transportation systems: A survey on forecasting tasks},
  author={Zhang, Zijian and Sun, Yujie and Wang, Zepu and Nie, Yuqi and Ma, Xiaobo and Li, Ruolin and Sun, Peng and Ban, Xuegang},
  journal={Transportation Research Record},
  pages={03611981251367699},
  year={2024},
  publisher={SAGE Publications Sage CA: Los Angeles, CA}
}

@article{chen2025scalable,
  title={Scalable Prediction of Heterogeneous Traffic Flow with Enhanced Non-Periodic Feature Modeling},
  author={Chen, Jing and Zhang, Sheng and Xu, Wenqiang},
  journal={Expert Systems with Applications},
  pages={128847},
  year={2025},
  publisher={Elsevier}
}

@article{wu2019graph,
  title={Graph wavenet for deep spatial-temporal graph modeling},
  author={Wu, Zonghan and Pan, Shirui and Long, Guodong and Jiang, Jing and Zhang, Chengqi},
  journal={arXiv preprint arXiv:1906.00121},
  year={2019}
}

@article{guo2024towards,
  title={Towards explainable traffic flow prediction with large language models},
  author={Guo, Xusen and Zhang, Qiming and Jiang, Junyue and Peng, Mingxing and Zhu, Meixin and Yang, Hao Frank},
  journal={Communications in Transportation Research},
  volume={4},
  pages={100150},
  year={2024},
  publisher={Elsevier}
}

@inproceedings{shao2025towards,
  title={Towards Trajectory Anomaly Detection: a Fine-Grained and Noise-Resilient Framework},
  author={Shao, Wei and Fang, Ziquan and Chen, Lu and Gao, Yunjun},
  booktitle={Proceedings of the 31st ACM SIGKDD Conference on Knowledge Discovery and Data Mining V. 2},
  pages={2490--2501},
  year={2025}
}

@article{liCausalInterventionWhat2025,
  title = {Causal Intervention Is What Large Language Models Need for Spatio-Temporal Forecasting},
  author = {Li, Shijie and Li, He and Li, Xiaojing and Xu, Yong and Lin, Zhenhong and Jiang, Huaiguang},
  date = {2025},
  journaltitle = {IEEE Transactions on Cybernetics},
  shortjournal = {IEEE Trans. Cybern.},
  pages = {1--13},
  issn = {2168-2275},
  doi = {10.1109/TCYB.2025.3569333},
  url = {https://ieeexplore.ieee.org/abstract/document/11017752},
}

@article{ni2016forecasting,
  title={Forecasting the subway passenger flow under event occurrences with social media},
  author={Ni, Ming and He, Qing and Gao, Jing},
  journal={IEEE Transactions on Intelligent Transportation Systems},
  volume={18},
  number={6},
  pages={1623--1632},
  year={2016},
  publisher={IEEE}
}

@article{ke2024interpretable,
  title={Interpretable mixture of experts for time series prediction under recurrent and non-recurrent conditions},
  author={Ke, Zemian and Duan, Haocheng and Qian, Sean},
  journal={arXiv preprint arXiv:2409.03282},
  year={2024}
}

@inproceedings{guo2019attention,
  title={Attention based spatial-temporal graph convolutional networks for traffic flow forecasting},
  author={Guo, Shengnan and Lin, Youfang and Feng, Ning and Song, Chao and Wan, Huaiyu},
  booktitle={Proceedings of the AAAI conference on artificial intelligence},
  volume={33},
  number={01},
  pages={922--929},
  year={2019}
}

@article{caliendo2008some,
  title={Some practical guidance for the implementation of propensity score matching},
  author={Caliendo, Marco and Kopeinig, Sabine},
  journal={Journal of economic surveys},
  volume={22},
  number={1},
  pages={31--72},
  year={2008},
  publisher={Wiley Online Library}
}

@inproceedings{li2018dcrnn,
  title={Diffusion convolutional recurrent neural network: Data-driven traffic forecasting},
  author={Li, Yaguang and Yu, Rose and Shah, Cyrus and Wang, Zihan},
  booktitle={International Conference on Learning Representations (ICLR)},
  year={2018}
}

\end{document}